\documentclass{article} 
\usepackage{iclr2016_conference,times}
\usepackage{hyperref}
\usepackage{url}

\usepackage{natbib}
\usepackage{graphicx}
\usepackage{amsmath}
\usepackage{amsfonts}
\usepackage{amssymb}
\usepackage{comment}

\usepackage{amsthm}

\newtheorem{theorem}{Theorem}[section]
\newtheorem{lemma}[theorem]{Lemma}

\theoremstyle{definition}
\newtheorem{definition}{Definition}[section]

\theoremstyle{remark}

\usepackage{wrapfig}

\title{Unitary-Group Invariant Kernels and Features from Transformed Unlabeled Data}

\author{Dipan K.~Pal \& Marios Savvides 
\\
Carnegie Mellon University\\
Pittsburgh, PA 15213 \\
\texttt{\{dipanp,msavvid\}@andrew.cmu.edu} \\
}

\begin{document}

\maketitle

\begin{abstract}
The study of representations invariant to common transformations of the data is important to learning. Most techniques have focused on local approximate invariance implemented within expensive optimization frameworks lacking explicit theoretical guarantees. In this paper, we study kernels that are invariant to the unitary group while having theoretical guarantees in addressing practical issues such as (1) unavailability of transformed versions of \textit{labelled} data and (2) not observing all transformations. We present a theoretically motivated alternate approach to the invariant kernel SVM. Unlike previous approaches to the invariant SVM, the proposed formulation solves both issues mentioned. We also present a kernel extension of a recent technique to extract linear unitary-group invariant features addressing both issues and extend some guarantees regarding invariance and stability.  We present experiments on the UCI ML datasets to illustrate and validate our methods.
\end{abstract}

\section{Introduction}\label{section_1}

It is becoming increasingly important to learn well generalizing representations that are invariant to many common transformations of the data. These transformations can give rise to many `degrees of freedom' even in a constrained task such as face recognition (\emph{e.g.} pose, age-variation, illumination etc.). In fact, explicitly factoring them out leads to improvements in recognition performance as found in \cite{leibo2014subtasks, hinton1987learning}. To this end, the study of invariant features is important.  \cite{AnselmiLRMTP13} showed that features that are explicitly invariant to intra-class transformations allow the sample complexity of the recognition problem to be reduced. 


\textbf{Prior Art: Invariant Kernels. }Kernel methods in machine learning have long been studied to considerable depth. Nonetheless, the study of invariant kernels and techniques to extract invariant features has received less attention. An invariant kernel allows the kernel product to remain invariant under transformations of the inputs. There has been some work on incorporating invariances into popular machinery such as the SVM in  \cite{lauer2008incorporating}. Most instances of incorporating invariances focused on \textit{local} invariances through regularization and optimization such as \cite{Schölkopf96incorporatinginvariances, scholkopf1998, decoste_2002, zhang2013learning}. Some other techniques were jittering kernels (\cite{scholkopf2002learning, decoste_2002}) and tangent-distance kernels (\cite{Haasdonk_tangent_distance}), both of which sacrificed the positive semi-definite property of its kernels and were computationally expensive. \cite{Haasdonk07invariantkernel} had first used group integration to arrive at invariant kernels, however, their approach does not address two important problems that arise in practice (group observation through unlabelled samples and partially observed groups). We will shortly state these problems more concretely and will show that the invariant kernels proposed do in fact solve both problems.

\textbf{Prior Art: Invariance through dataset augmentation. }Many approaches in the past have enforced invariance through generating transformed \textit{training} samples in some form such as \cite{Poggio92recognitionand, scholkopf2002learning, scholkopf1998, Niyogi98incorporatingprior, Reisert_2008, Haasdonk07invariantkernel}. This assumes that one has knowledge about the transformation. The approach presented in this paper however, under the unitarity assumption, can learn the transformations through unlabelled samples and does not need training dataset augmentation. Perhaps, the most popular method for incorporating invariances in SVMs is the virtual support method (VSV) in \cite{Schölkopf96incorporatinginvariances}, which used sequential runs of SVMs in order to find and augment the support vectors with transformed versions of themselves.  \cite{loosli_2007} proposed a similar algorithm to generate and prune out examples. Though these methods have had some success, most of them still lack explicit theoretical guarantees towards invariance. The proposed invariant kernel SVM formulation on the other hand, is guaranteed to be invariant. Further, unlike VSV and other approaches to incorporate invariance into the SVM, the proposed invariant kernel SVM solves the common and important practical problems that we will state shortly. To the best of our knowledge, it is the first formulation to do so.

\textbf{Prior Art: Linear Invariant Features. }  Recently, \cite{ AnselmiLRMTP13} proposed linear group-invariant features as an explanation for multiple characteristics of the visual cortex. They achieve invariance in a slightly more general way than group integration, utilizing measures of the distribution characterizing an orbit of a sample under the action group. We extend the method to the RKHS using unitary kernels and extend some properties regarding invariance and stability. We also show that the extension can solve both motivating problems (Problem 1 and Problem 2). This leads to a practical way of extracting non-linear invariant features with theoretical guarantees.

\textbf{Motivating Problems. } We now state the two central problems that this paper tries to address through invariant kernels and features. A common practical problem that one faces utilizing previous methods involving generating transformed samples is the computational expense of generating and processing them (including virtual support vectors). Further, in many cases transformed  \textit{labelled} samples are unavailable.Two important problems that arise when practically applying invariant kernels and features are:

\textit{\textbf{Problem 1: (Group observed through unlabelled samples)} The transformed versions of the training labelled data are not available \emph{i.e.} one might only have access to transformed versions of unlabelled data outside of the training set (theoretically equivalent to having transformed versions of arbitrary vectors), \emph{e.g.} only unlabelled transformed images are observed.}

\textit{\textbf{Problem 2: (Partially observed group)} Not all members of the group (symmetric set) of transformations $\mathcal{G}$ are observed \emph{i.e.} the group is only partially observed through its actions, \emph{e.g.} not all transformations of an image are observed. In many practical cases, partial invariance is in fact necessary, when a transformation from one class to another exists.}




\textbf{Group Theory and Invariance. }Towards this goal, the study of incorporating invariance through group integration seems useful. Group theory is an elegant way to model symmetry. Classical invariant theory provides group integration techniques to enforce invariance.  Group integration can also be used to model mean pooling (and max pooling albeit in a different framework as proposed in \cite{AnselmiLRMTP13}), which is in implicit use in several areas of  machine learning and computer vision. The transformations, in this paper, are modelled as \textit{unitary} and collectively form the unitary-group $\mathcal{G}$. Classes of learning problems, such as vision, often have transformations belonging to the unitary-group, that one would like to be invariant towards (such as translation and rotation). The results can also be extended to discrete groups. In practice however, \cite{liao2013} found that invariance to much more general transformations not captured by this model can been achieved.  

We will see that given explicit access to $\mathcal{G}$, one can theoretically capitalize on properties such as guaranteed global invariance (as opposed to local invariance in optimization approaches, where the classifier is invariant to only small transformations). However, controlled \textit{local} invariance can also be achieved. Local invariance is important when the extreme transformation of one class overlaps with another. The unitary property of the group and the unitary restriction on kernels (in Section~\ref{sec_group_int_rkhs}), allow the development of theoretical motivation for existing techniques, an invariant kernel and invariant kernel features theoretically addressing Problems 1 and 2.

\textbf{Contributions. } We list our main contributions below:
\begin{enumerate}
\item In contrast to many previous studies on invariant kernels, our focus is to study positive semi-definite unitary-group invariant kernels and features guaranteeing invariance that can address \textit{both} Problem 1 and  Problem 2 .

\item One of our central results to applying group integration in the RKHS builds on the observation that, under unitary restrictions on the kernel map, group action is preserved in the RKHS.

 \item Using the proposed invariant kernel, we present a theoretically motivated alternate approach to designing a non-linear invariant SVM that can handle both Problem 1 and Problem 2 with explicit invariance guarantees.

 \item We propose kernel unitary-group invariant feature extraction techniques by extending a theory of linear group invariant features presented in \cite{AnselmiLRMTP13}. We show that the kernel extension addresses both Problem 1 and Problem 2 and preserves properties such as global (and local) invariance and stability.
\end{enumerate}

\textbf{Organization. } The paper is broadly organized into two parts. Section~\ref{sec_group_invariance} and \ref{sec_inv_kernel} present the proposed invariant kernels and the invariant kernel SVM, whereas Section~\ref{section_global_inv_magictheory} and \ref{section_partial_inv_magictheory} present the proposed invariant features extracted using kernels. 

\textbf{Section~\ref{sec_group_invariance} and \ref{sec_inv_kernel} (Unitary-group Invariant Kernels). }We first present some important known elementary unitary-group integration properties and present a central result to applying group integration in the RKHS in Section~\ref{sec_group_invariance}. We then present a theoretically motivated alternate approach to designing a non-linear invariant SVM and present a simple albeit important result to reduce computation.  In Section~\ref{sec_inv_kernel}, we then continue on to develop an invariant kernel which does not require to observe transformed versions of the input arguments whatsoever. 

\textbf{Section~\ref{section_global_inv_magictheory} and \ref{section_partial_inv_magictheory} (Unitary-group Invariant Kernel Features). }In Section~\ref{section_global_inv_magictheory}, we propose kernel unitary-group invariant feature extraction techniques by extending a linear invariant feature extraction method (\cite{AnselmiLRMTP13}) to the kernel domain. We show that the resultant feature, while addressing Problem 1, preserves important properties such as global invariance and stability. In Section~\ref{section_partial_inv_magictheory}, we show that a simple extension of the method can help it to solve both problems (Problem 1 and Problem 2). This leads to a practical way of extracting invariant non-linear features with theoretical guarantees.

\textbf{Section~\ref{sec_exp}} presents some experiments illustrating our methods. 

\section{Globally Group Invariant Kernels: When the group $\mathcal{G}$ is explicitly known}\label{sec_group_invariance}

\textbf{Premise:} Consider a dataset of normalized samples along with labels $\mathcal{X} = \{x_i, y_i\}~ \forall i\in 1...N$ with $x \in \mathbb{R}^d$ and $y\in \{+1,-1\}$. We now introduce into the dataset a number of unitary transformations $g$ part of the locally compact unitary-group $\mathcal{G}$ (in general we require local compactness to enable the existence of the Haar measure). Our augmented normalized dataset becomes $\mathcal{X}_\mathcal{G} = \{ gx_i, y_i\}~ \forall g \in \mathcal{G}~\forall i$\footnote{With a slight abuse of notation, we denote by $gx$ the action of group element $g\in\mathcal{G}$ on $x$}. Thus, $\mathcal{X} \subseteq \mathcal{X}_\mathcal{G} $. We assume for now that $\mathcal{G}$ is known and accessible completely. Let $\phi$ be some mapping to a high dimensional Hilbert space $\mathcal{H}$, \emph{i.e.}$~\phi: \mathcal{X} \rightarrow \mathcal{H}$. Once the points are mapped, the problem of learning a separator in that space can be assumed to be linear. 



An invariant function is defined as follows.

\begin{definition}[\textit{$\mathcal{G}$-Invariant Function}]\label{def_invariant}
For any group $\mathcal{G}$, we define a function $f: \mathcal{X} \rightarrow \mathbb{R}^n$ to be $\mathcal{G}$-invariant if $f(x) = f(gx)~\forall x\in \mathcal{X} ~\forall g\in \mathcal{G}$.
\end{definition}

One method of generating an invariant towards a group is through group integration. Group integration has stemmed from  classical invariant theory and its foundational theorem was proved by Haar.

\begin{theorem} (Haar) On every locally compact group there exists at least one left invariant integral. Such an integral is unique except for a strictly positive factor of proportionality.\end{theorem}

One can choose the factor of proportionality such that the group volume equals 1  (\emph{i.e.} $\int 1 dg =1$, in the case of discrete finite groups each group element would be scaled down by $\frac{1}{|\mathcal{G}|}$). For compact groups, such an integral converges for any bounded function of the group. For discrete groups, the integral is replaced by a sum. Group integration can be shown to be a projection onto a $\mathcal{G}$-invariant subspace. Such a subspace can be defined for a Hilbert space $\mathcal{H}$ by $\mathcal{H}^\mathcal{G} = \{ x \in \mathcal{H} ~|~ x=gx ~\forall g\in \mathcal{G},~\forall x\in \mathcal{H} \}$. An invariant to any group $\mathcal{G}$ can be generated through the following basic (previously) known property (Lemma~\ref{lem_invariance}) based on group integration.


\begin{lemma}\label{lem_invariance} (Invariance Property) Given a vector $\omega\in \mathbb{R}^d$, and any group $\mathcal{G}$, for any fixed $ g' \in \mathcal{G}$ and a normalized Haar measure $dg$, we have $g'\int_\mathcal{G} g \omega ~dg = \int_\mathcal{G} g \omega ~dg$
\end{lemma}


The Haar measure ($dg$) exists for every locally compact group and is unique upto a positive multiplicative constant (hence normalized). A similar property holds for discrete groups. The Invariance Property results in global invariance to group $\mathcal{G}$. This property allows one to generate a $\mathcal{G}$-invariant subspace in the inherent space $\mathbb{R}^N$. 

The following two lemmas (Lemma~\ref{transpose} and \ref{unitary_projection}) showcase (novel) elementary properties of the operator $\Psi = \int_\mathcal{G} g~dg$ for the unitary-group $\mathcal{G}$. These properties would prove useful in the analysis of unitary-group invariant kernels and features.

\begin{lemma} \label{transpose} If $\Psi = \int_\mathcal{G} g~dg$ for unitary $\mathcal{G}$, then $\Psi^T = \Psi$
\end{lemma}


\begin{lemma}\label{unitary_projection} (Unitary Projection) If $\Psi = \int_\mathcal{G} g~dg$ for any $\mathcal{G}$, then $\Psi\Psi = \Psi$, \emph{i.e.} it is a projection operator. Further, if $\mathcal{G}$ is unitary, then $\langle \omega, \Psi \omega' \rangle = \langle \Psi \omega, \omega'  \rangle~\forall \omega, \omega' \in \mathbb{R}^d$
\end{lemma}




The proofs of these lemmas utilize elementary properties of groups, invariance of the Haar measure $dg$ and the unitarity of $g$\footnote{All proofs are presented in the supplementary material}. 

\textbf{Sample Complexity and Generalization. }On applying the operator $\Psi$ to the dataset $\mathcal{X}$, all points in the set $\{ gx ~|~ g \in \mathcal{G} \}$ for any $x \in \mathcal{X}$ map to the same point $\Psi x$ in the $\mathcal{G}$-invariant subspace. Theoretically, this would drastically reduce sample complexity while preserving linear feasibility (separability). It is trivial to observe that  \textit{a perfect linear separator learnt in $\mathcal{X}_\Psi = \{ \Psi x~ |~ x \in \mathcal{X} \}$ would also be a perfect separator for $\mathcal{X}_\mathcal{G}$}, thus in theory achieving perfect generalization. We prove a similar result for the RKHS case in Section~\ref{sec_inv_svm}. This property is theoretically powerful since cardinality of $\mathcal{G}$ can be large. A classifier can avoid having to observe transformed versions $\{gx\}$ of any $x$ and yet generalize.

\subsection{Group Actions Reciprocate in a Reproducing Kernel Hilbert Space}\label{sec_group_int_rkhs}

Group integration provides exact invariance in the domain of $\mathcal{X}$. However, it requires the group structure to be preserved. In the context of kernels, it is imperative that the group relation between the samples $x\in \mathcal{X}_\mathcal{G}$ be preserved in the kernel Hilbert space $\mathcal{H}$ corresponding to some kernel $k$. Under the restriction of unitary $k$, this is possible. We present an elementary albeit important result that allows this after defining unitary kernels in the following sense.

\begin{definition}[\textit{Unitary Kernel}]\label{def_unitary_kernel}
We define a kernel $k(x, y) = \langle \phi(x), \phi(y) \rangle$ to be a unitary kernel if, for a unitary group $\mathcal{G}$, the mapping $\phi(x): \mathcal{X} \rightarrow \mathcal{H}$ satisfies $\langle \phi(gx), \phi(gy)\rangle = \langle \phi(x), \phi(y)\rangle~\forall g\in \mathcal{G}, \forall x, y \in \mathcal{X}$.
\end{definition}
The unitary condition is fairly general, a common class of unitary kernels is the RBF kernel. We now define an operator $g_\mathcal{H}:\phi(x) \rightarrow \phi(gx)~~\forall \phi(x) \in \mathcal{H} $ for any $g\in \mathcal{G}$ where $\mathcal{G}$ is unitary. $g_\mathcal{H}$ thus is a mapping within the RKHS. Under unitary $\mathcal{G}$, we then have the following result.
\begin{theorem}\label{theorem_1}
(Covariance in the RKHS) If $k(x, y) = \langle  \phi(x), \phi(y)  \rangle$ is a unitary kernel in the sense of Definition~\ref{def_unitary_kernel}, then  $g_\mathcal{H}$ is unitary, and the set $\mathcal{G}_\mathcal{H} = \{ g_\mathcal{H} ~|~ g_\mathcal{H}: \phi(x)\rightarrow \phi(gx)~\forall g\in \mathcal{G} \}$ is a unitary-group in $\mathcal{H}$.
\end{theorem}



Theorem~\ref{theorem_1} shows that the unitary-group structure is preserved in the RKHS. This provides new theoretically motivated approaches to achieve invariance in the RKHS. Specifically, a theory of invariance which was proposed to utilize unsupervised linear filters can now also utilize non-linear supervised `templates' as we discuss in Section~\ref{section_global_inv_magictheory}.



\subsection{Invariant Non-linear SVM: An Alternate Approach Through Group Integration}\label{sec_inv_svm}


We present the group integration approach to kernel SVMs before comparing it to other methods. The decision function of SVMs can be written in the general form as $f_\theta(x) = \omega^T \phi(x) + b$ for some bias $b\in \mathbb{R}$ (we agglomerate all parameters of $f$ in $\theta$) where $\phi$ is the feature map, \emph{i.e.}$~\phi: \mathcal{X} \rightarrow \mathcal{H}$. Reviewing the SVM, a maximum margin separator is found by minimizing loss functions such as the hinge loss along with a regularizer. In order to invoke invariance, we can now utilize group integration in the the kernel space $\mathcal{H}$ using Theorem~\ref{theorem_1}. All points in the set $\{ gx \in \mathcal{X}_\mathcal{G}\}$ get mapped to $\phi(gx) = g_\mathcal{H}\phi(x)$ for a given $g\in \mathcal{G}$. Group integration then results in a $\mathcal{G}$-invariant subspace within $\mathcal{H}$ through $\Psi_\mathcal{H} = \int_\mathcal{G_\mathcal{H}} g_\mathcal{H}~dg_\mathcal{H}$ using Lemma~\ref{lem_invariance}. Introducing Lagrange multipliers $\alpha = (\alpha_1, \alpha_2 ...\alpha_N)\in \mathbb{R}^N$, the dual formulation (utilizing Lemma~\ref{transpose} and Lemma~\ref{unitary_projection}) then becomes 


\begin{align}\label{eq_svm}
\arg\min_\alpha -\sum_i \alpha_i + \frac{1}{2}\sum_i\sum_j y_i y_j \alpha_i \alpha_j \langle \Psi_\mathcal{H}\phi( x_i), \Psi_\mathcal{H}\phi(x_j) \rangle
\end{align}

under the constraints $\sum_i \alpha_i y_i = 0, ~~~ 0 \leq \alpha_i \leq \frac{1}{N} ~~\forall i$. The separator is then given by $\omega^*_\mathcal{H} = \sum_i y_i \alpha_i \Psi_\mathcal{H} \phi(x_i) = \Psi_\mathcal{H}\omega^*$ thereby existing in the $\mathcal{G}_\mathcal{H}$-invariant (or equivalently $\mathcal{G}$-invariant) subspace $\Psi_\mathcal{H}$ within $\mathcal{H}$ (since $g \rightarrow g_\mathcal{H}$ is a bijection). Effectively, the SVM observes samples from $\mathcal{X}_{\Psi_\mathcal{H}} = \{ x~|~ \phi(x)=\Psi_\mathcal{H}\phi(u), ~\forall u\in \mathcal{X}_\mathcal{G}\}$. If $\mathcal{G}$ is known, then this provides \textit{exact global} invariance during testing. Further, \textit{$\Psi_\mathcal{H}\omega^*$ is a maximum-margin separator of $\{\phi(\mathcal{X}_G)\}$}. This can be shown by the following result.

\begin{theorem}\label{th_generalization}(Generalization)  \textit{For a unitary group $\mathcal{G}$ and unitary kernel $k(x, y) = \langle \phi(x), \phi(y)  \rangle$, if $\omega^*_\mathcal{H} = \int_\mathcal{G_\mathcal{H}} g_\mathcal{H}~dg_\mathcal{H}~\omega^* = \Psi_\mathcal{H}\omega^*$ is a perfect separator for $\{\Psi_\mathcal{H}\phi(\mathcal{X})\} = \{ \Psi_\mathcal{H}\phi(x)~|~\forall x\in \mathcal{X} \}$, then $\Psi_\mathcal{H}\omega^*$ is also a perfect separator for $\{\phi(\mathcal{X}_G)\} = \{ \phi(x)~|~x\in \mathcal{X}_\mathcal{G} \}$ with the same margin. Further, a max-margin separator of $\{\Psi_\mathcal{H}\phi(\mathcal{X})\}$ is also a max-margin separator of $\{\phi(\mathcal{X}_G)\}$.}
\end{theorem}

The invariant non-linear SVM in objective~\ref{eq_svm}, observes samples in the form of $\Psi_\mathcal{H}\phi(x)$ and obtains a max-margin separator $\Psi_\mathcal{H}\omega^*$. Theorem~\ref{th_generalization} shows that the margins of $\phi(\mathcal{X}_\mathcal{G})$ and $\{\Psi_\mathcal{H}\phi(\mathcal{X})\}$ are deeply related and implies that $\Psi_\mathcal{H}\phi(x)$ is a max-margin separator for both datasets. Theoretically, the Invariant non-linear SVM is able to generalize to $\mathcal{X}_\mathcal{G}$ on \textit{just} observing $\mathcal{X}$ and utilizing prior information in the form of $\mathcal{G}$ for all unitary kernels $k$. This is true \textit{in practice} for linear kernels. For non-linear kernels in practice, however, the invariant SVM still needs to observe and integrate over transformed \textit{training} inputs. We also present the following result for unitary-group invariant kernels which helps in saving computation. 
\begin{lemma}\label{invariant_proj} (Invariant Projection) If $\Psi = \int_\mathcal{G} g~dg$ for any unitary group $\mathcal{G}$, then for any fixed $g'\in \mathcal{G}$ we have  $\langle \Psi\omega, \Psi \omega' \rangle = \langle g' \omega, \Psi\omega'  \rangle~\forall \omega, \omega' \in \mathbb{R}^d$
\end{lemma}
We provide the proof in the supplementary material. Thus, the kernel in the Invariant SVM formulation can be replaced by the form $k_\Psi(x, y) = \langle \phi(x), \Psi_\mathcal{H}\phi(y) \rangle $, thereby reducing the number of transformed training samples required to be observed by an order of magnitude. \textit{It also allows for the kernel $k_\Psi(x, y)$ to be invariant to the orbit of $x$, \emph{i.e.} $\{ gx \}$ with observing just a single arbitrary point ($g'x$) on the orbit.} Nonetheless, as the formulation stands, it still requires observing the entire orbit of atleast one of the transformed training samples. However, we can get around this fundamental problem as we show in the next section (Section~\ref{sec_inv_kernel}). 

 Note that for the general kernel, the $\mathcal{G}_\mathcal{H}$-invariant subspace cannot be explicitly computed, it is only implicitly projected upon through $\Psi_\mathcal{H}\phi(x_i) = \int_\mathcal{G}\phi(gx_i)dg_\mathcal{H}$. It is important to note that during \textit{testing} however, the SVM formulation will be invariant to transformations of the test sample regardless of a linear or non-linear kernel. Also, interestingly, $\omega^*$ might be a different decision boundary than $\omega^{'*}$ obtained by training the vanilla SVM on $\mathcal{X}_\mathcal{G}$.



\textbf{Positive Semi-Definiteness. }The $\mathcal{G}$-invariant kernel map is now of the form $k_\Psi(x, y) = \langle \phi(x), \int_\mathcal{G}\phi(gy)dg_\mathcal{H} \rangle$. \textit{This preserves the positive semi-definite property of the kernel $k$ while guaranteeing global invariance to unitary transformations.}, unlike jittering kernels (\cite{scholkopf2002learning, decoste_2002}) and tangent-distance kernels (\cite{Haasdonk_tangent_distance}). If we wish to include invariance to \textit{scaling} however, then we would lose positive-semi-definiteness (it is also not a unitary transform). Nonetheless, \cite{walder2007learning} show that conditionally positive definite kernels still exist for transformations including scaling, although we focus of unitary transformations in this paper. 

\textbf{Partial Invariance. }The invariant kernel SVM formulation (objective~\ref{eq_svm}) also supports partial invariance when $\mathcal{G}$ is not fully observed (addressing Problem 2), a notion extended to invariant kernel methods in Section~\ref{section_partial_inv_magictheory}. Partial invariance gives one control over the degree of invariance over transformation groups, allowing classes that are transformations of one another (such as MNIST classes $6$ and $9$) to be discriminated.

\textbf{Relating the Virtual Support Vector Method (VSV):} Consider the popular Virtual Support Vector Method (VSV) (\cite{Schölkopf96incorporatinginvariances}).  Here the support vectors are augmented with small (finite) number of transformed versions of themselves. \textit{This assumes that the transformations are explicitly known}, thereby failing to address Problem 1. The augmented training set is used to train another SVM with improved invariance.  We show in the following section that the Invariant SVM formulation (objective~\ref{eq_svm}), on the other hand, does address Problem 1. The group integration framework provides a theoretical motivation for the VSV, since at minimum, it suggests having transformed versions of the support vectors. The VSV however, can have different $\alpha_i$ for different transformed versions of a $x_i$, whereas group integration would force them to be the same because the kernel $k_\Psi(x, y)$ is $\mathcal{G}$-invariant. For \textit{linear kernels} we have more benefits. Group integration also suggests building an explicit $\mathcal{G}$-invariant subspace before projecting the training set on it. This approach does not increase computation time (for \textit{linear kernels}) while allowing the SVM to generalize to $\mathcal{G}$-transformed inputs. 


\section{Globally Group Invariant Kernels: When action of group $\mathcal{G}$ is observed \textit{only} on \textit{unlabelled} data} \label{sec_inv_kernel}

The previous section introduced a group integration approach to the invariant non-linear SVM. Although the formulation addresses Problem 2, it does not address Problem 1 \emph{i.e.} the kernel $k_\Psi(x, y) = \langle \int_\mathcal{G}\phi(gx)dg_\mathcal{H}, \int_\mathcal{G}\phi(gy)dg_\mathcal{H} \rangle = \langle \Psi_\mathcal{H}\phi(x), \Psi_\mathcal{H}\phi(y) \rangle$ still requires observing transformed versions of the \textit{labelled} input sample namely $\{gx ~|~ gx\in \mathcal{X}_\mathcal{G} \}$ (or atleast one of the labelled samples if we utilize Lemma~\ref{invariant_proj}). We now present an approach to not require the observation of any \textit{labelled} training sample whatsoever.

Assume that for every sample $x\in\mathcal{X}_\mathcal{G}$, there exists a vector $u_x$ s.t. $\phi(x) \approx \phi(T)u_x$, where $T$ is an arbitrary unlabelled set (in the form of a column-major matrix) of $M$ arbitrary templates $\{ t_i \}$ (Note that there exist more informed ways of choosing $T$, however to keep the theory general we work with arbitrary template sets). We assume that we have access to transformed versions of each template $t_i$ \emph{i.e.} we observe $\mathcal{G}$ \textit{only} through $\{ gt ~|~  t\in T, g\in\mathcal{G}\}$. We then have the following result.

\begin{theorem}\label{invariant_kernel}
For a unitary group $\mathcal{G}$, a template set $T  \in \mathbb{R}^{d\times M} = \{t_i\}$ and a unitary kernel $k(x, y) = \langle \phi(x), \phi(y) \rangle$, if  $\phi(x)= \phi(T)u_x$ and  $\phi(y)=\phi(T)u_y$, then the $\mathcal{G}$-invariant kernel $k_\Psi(x, y) = \langle \Psi_\mathcal{H}\phi(x), \Psi_\mathcal{H}\phi(y) \rangle$ can be written as $$k_\Psi(x, y) = \int_\mathcal{G} \langle \phi(gT) u_x , \phi(T)u_y \rangle dg_\mathcal{H}$$
\end{theorem}

Theorem~\ref{invariant_kernel} assumes that the points $\phi(x)$ lies in the span of $\phi(T)$. It allows the kernel $k_\Psi(x, y)$ to be $\mathcal{G}$-invariant for $x, y$ \emph{i.e.} $k_\Psi(x, y) = k_\Psi(g'x, g''y)~\forall g', g'' \in \mathcal{G}$. \textit{It achieves this while only observing transformed versions of the unlabelled template set $T$}. This is very useful since the use of Theorem~\ref{invariant_kernel} solves Problem 1 while guaranteeing invariance. Further, in practice, one does not need to have explicit knowledge of the transformations. In many cases, they can simply store the naturally transforming samples (\emph{e.g.} transforming images). A constructed kernel can be applied to any dataset directly provided the same group $\mathcal{G}$ acts. Coefficients $u_x$ required for Theorem~\ref{invariant_kernel} for any $x\in\mathcal{X}_\mathcal{G}$ can be approximated by projecting the sample $x$ onto the space spanned by $T$ in the RKHS \emph{i.e.} $u_x = (\phi(T)^T\phi(T))^{-1}\phi(T)^T\phi(x)$. This assumes that the kernel matrix $(\phi(T)^T\phi(T))^{-1}$ is invertible, a condition that can be satisfied by construction.

\textbf{Invariant Non-linear SVM through transformed unlabelled data (comparison with the VSV):} The invariant kernel SVM in objective~\ref{eq_svm} using the invariant kernel $k_\Psi(x, y)$ achieves invariance through learning the transformation \textit{only} through observed unlabelled data. Further, it does not need multiple runs as opposed to the VSV which requires the generation of transformed labelled examples. Theorem~\ref{invariant_kernel} allows an invariant kernel to be used \textit{directly} without the computational expense of finding potential support vectors, generating transformations of them and then processing the added samples. Further, invariance helps to reduce sample complexity and improve performance given a number of samples, a phenomenon we observe in our experiments.

\section{Globally Group Invariant Kernel Features from a Single Sample: When action of $\mathcal{G}$ is observed \textit{only} on unlabelled data} \label{section_global_inv_magictheory}


Up until now we have studied the properties of the proposed unitary group invariant kernels. We now shift our attention to group invariant \textit{features}. Invariant kernels are a form of an invariant similarity measure and can be used to construct invariant feature maps. \cite{ AnselmiLRMTP13} proposed linear invariant features that enjoys properties such as global invariance and stability. We extend their method to the RKHS using unitary kernels and extend the invariance and stability properties. We now briefly present their theory of invariance.

\subsection{Theory of Linear Invariant Features}

Under $\mathcal{G}$, the orbit of any sample $x \in \mathcal{X}$ is defined by $\mathcal{O}_x = \{ gx~|~ g\in \mathcal{G} \}$. As a straightforward albeit elegant observation, \textit{the orbit itself is an invariant under $\mathcal{G}$, since $\mathcal{O}_x = \mathcal{O}_{gx}$}. Measures of such an orbit also provide invariance, such as the high dimensional distribution $P_x$ induced by the group's action on $x$. In fact, \cite{ AnselmiLRMTP13} show $P_x$ to be both invariant and unique, \emph{i.e.} $x \thicksim x' \Leftrightarrow \mathcal{O}_x \thicksim \mathcal{O}_{x'} \Leftrightarrow P_x \thicksim P_{x'}$, where $\thicksim$ denotes membership in the same class. Thus, measures of the distribution, through a finite number of one-dimensional projections $\{ P_{\langle x, t^k \rangle} \}^K_{k=1}$, can be used as a similarity measure between two orbits \footnote{This follows from Cramer-Wold theorem along with concentration of measures.}. Further, the measures are invariant to the action of unitary group $\mathcal{G}$. For \textit{unitary} group $\mathcal{G}$, normalized dot-products and an \textit{arbitrary} template $t^k$, an empirical estimate of the 1-dimensional distribution of the projection onto template $t^k$ can be expressed as $\mu^k_n(x) =  \int_\mathcal{G} \eta_n (\langle x, g  t^k \rangle)dg \label{eq_pooling}$, where $\eta_n~\forall n\in \{1...N\}$, a non-linearity, can either estimate the $n$-th bin of the CDF or the $n$-th moment, the set of which together define $P_{\langle x, t^k \rangle}$. In practice, \cite{liao2013} found that a few or even one of these moments has been shown to be sufficiently invariant. The final signature or feature vector is $\Delta(x) = (\{ \mu^1_n(x) \}, \{ \mu^2_n(x) \} ... \{ \mu^K_n(x) \}, ) \in \mathbb{R}^{NK} ~\forall n$.





\subsection{Group Invariant Feature Extraction in Kernel Space from a \textit{Single} Sample}

We now present a kernel extension of the approach to invariance presented above.  We assume access to the set $~\mathcal{U} = \{ \mathcal{O}_{t^k}~|~ \forall k \in \{1...K\} \}$, \emph{i.e.} the orbits of $K$ arbitrary unlabelled vectors or templates. For simplicity, we also assume a compact unitary $\mathcal{G}$ with finite cardinality $|\mathcal{G}|$. Then for every $g'\in \mathcal{G}$, we have template $t^k_{g'} = g' t^k$. Similarly, for unitary kernels (Definition~\ref{def_unitary_kernel}), the templates in the RKHS behave as transformed versions of each other owing to Theorem~\ref{theorem_1}. Therefore, $t^k_{\mathcal{H} g'} = \phi(t^k_{g'}) = g'_\mathcal{H} \phi(t^k)~\forall g' \in \mathcal{G}$. Thus,  $\{ g\phi(t^k)~|~g\in \mathcal{G} \}$ form a set of transformed elements for each $k$ under the action of $\mathcal{G}$. Invariance can then be achieved using a form of Equation 7 in \cite{AnselmiLRMTP13}.

\begin{align}\label{non_linear_m_theory}
\Upsilon^k_n(x) = \frac{1}{|\mathcal{G}|} \sum_g \eta_n (\langle \phi(x),  t^k_{\mathcal{H}g} \rangle)
\end{align}

$\Upsilon(x)$ can extract non-linear kernel features for any \textit{single} sample $x\in\mathcal{X}$ that are invariant to the group $\mathcal{G}$ without ever needing to observe $\{gx ~|~ g\in\mathcal{G}\setminus I\}$ \footnote{Note that even though the features extracted are non-linear, invariance generated is purely towards unitary transformations.}.  This also solves Problem 1 listed in the introduction. Recall that $\eta_n$ can either estimate the CDF or the set of moments. In the case of moments, the first moment leads to \textit{mean pooling} and the infinite moment results in \textit{max pooling}. We now show that the kernel feature $\Upsilon(x)$ continues to satisfy useful properties such as stability (in the Lipschitz sense) \emph{i.e.} a form of a stability result in \cite{AnselmiLRMTP13} can be proved using a similar analysis. 

\begin{theorem}\label{theorem_stability}
(Stability) If  $\Upsilon(x)$ is invariant to a unitary-group $\mathcal{G}$ and the non-linearities $\eta_n$ are Lipschitz continuous with constant $L_{\eta_n}$, with $L_\eta = \max_n(L_{\eta_n}) ~s.t.~ NL_\eta\leq \frac{1}{\sqrt{2}}$, and we have a normalized unitary kernel $k$ with $k(x, x)=1, \forall x\in \mathbb{R}^d$, then $$\|\Upsilon(x) - \Upsilon(x')\|^2_2 < 1 - k(x, x')_H  \leq 1 - k(x, x')$$ for all $x, x' \in \mathcal{X}$. Here $k(x, x')_H$ is the kernel distance in the Hausdorff sense in $\mathcal{H}$, \emph{i.e.} $k(x, x')_H = \max_{g, g' \in\mathcal{G}}k(gx, g'x')$.
\end{theorem}


A good representation ideally should be stable and the distance between two points in the feature space should be bounded. Unstable representations can skew the feature space and allow for degenerate results. Theorem~\ref{theorem_stability} shows that under Lipschitz continuity for the $\eta_n$ estimation functions and $k(x, x)=1, \forall x\in \mathbb{R}^d$, the kernel feature distance is bounded by the kernel product.

\textbf{Discriminative templates:} Equation~\ref{non_linear_m_theory} can be instantiated to extract \textit{discriminative} kernel features, by choosing discriminative instead of arbitrary templates. Let $\mathcal{U}_\mathcal{H} = \phi(\mathcal{U})$, then for each group element $g'\in \mathcal{G}$, one can train $K$ binary one \emph{vs.} all classifiers with the $k^{th}$ template ($g't^k$) labelled as $+1$ and the rest ($\{ gt^k ~|~g\in \mathcal{G} \setminus\{ g'\} \}$) as $-1$ for all $k$. Recall that the separator $\omega_{g'}^k$ (for template $t^k$ as $+1$ and for  group element $g'$) can be expressed $\omega_{g'}^k = \sum_i \alpha_i y_i \phi(g't^k) = g'_\mathcal{H}( \sum_i \alpha_i y_i \phi(t^k)) = g'_\mathcal{H}\omega_{I}^k$ (using Theorem~\ref{theorem_1} and where $I$ is the identity element of $\mathcal{G}$). Thus,  $\{ \omega_{I}^k, ...,\omega_{g'}^k \}$ form a set of transformed templates for each $k$ under the action of $\mathcal{G}$ using which partial invariance can then be achieved through Equation~\ref{non_linear_m_theory}\footnote{Since this is agnostic to the selection of $\alpha$, any classifier which can be expressed as a linear combination of the samples in $\mathcal{H}$ (such as the perceptron, SVM, correlation filters) can be used as discriminative templates to generate invariance.}. 


\section{Towards Partially Group Invariant Kernels: When the group $\mathcal{G}$ is \textit{partially} observed through transformed samples}\label{section_partial_inv_magictheory}

We extend the notion of partial invariance to the kernel features extracted similar to Equation~\ref{non_linear_m_theory} following the analysis of \cite{AnselmiLRMTP13}. Partial invariance arises from partially observing the group $\mathcal{G}$, \emph{i.e.} observing only a finite group (may not be a subgroup) $\mathcal{G}_0\subseteq \mathcal{G}$. In practice, this is the most likely case. However, partial invariance can be obtained over the observed subset $\mathcal{G}_0$ through a local kernel feature, which can also be generalized to locally compact groups. A partially invariant kernel feature is $\widehat{\Upsilon}^k_n(x) = \frac{1}{|\mathcal{G}_0|} \sum_{g\in\mathcal{G}_0} \eta_n (\langle \phi(x),  \omega_g^k \rangle)$.


\textbf{Uniqueness:} The analysis for uniqueness in \cite{AnselmiLRMTP13} can be applied to $\widehat{\Upsilon}(x)$ with no significant changes, since the group structure is preserved in $\mathcal{H}$ through Theorem~\ref{theorem_1}. In summary, \textit{any two partial orbits with a common point are identical}.\\
\textbf{Invariance:} Theorem 6 from \cite{AnselmiLRMTP13} can be applied in $\mathcal{H}$ with some modification.
\begin{theorem}\label{theorem_partial_inv}
 (Partially Invariance) Let $\eta_n: \mathbb{R}\rightarrow \mathbb{R}^+$ are a set of bijective and positive functions and $\mathcal{G}$ be a locally compact group. Further, assuming $\mathcal{G}_0\subseteq\mathcal{G}$ and  $supp(\langle gx, \omega^k_g \rangle) \subseteq \mathcal{G}_0$ (where supp() denotes the support), then $\forall \overline{g}\in\mathcal{G}$ and $\forall x\in \mathbb{R}^d$, we have $\widehat{\Upsilon}^k_n(x) = \widehat{\Upsilon}^k_n(\overline{g}x)$.
\end{theorem}
\textbf{Stability:} $\widehat{\Upsilon}(x)$ is stable (in the Lipschitz sense) following the analysis of Theorem~\ref{theorem_stability}. In particular, we have the following result. 
\begin{theorem}\label{theorem_stability_partial}
 (Stability of Partially Invariant Feature) If $\widehat{\Upsilon}(x)$ is partially invariant to the group $\mathcal{G}_0$ and the non-linearities $\eta_n$ are Lipschitz continuous with constant $L_{\eta_n}$, with $L_\eta = \max_n(L_{\eta_n}) ~s.t.~ NL_\eta\leq \frac{1}{\sqrt{2}}$, and we have kernel $k$ with $k(x, x)=1, \forall x\in \mathbb{R}^d$, then for unitary $\mathcal{G}$, if $\mathcal{G}_0\subseteq \mathcal{G}$ and  assuming $supp(\langle gx, \omega^k_g \rangle) \subseteq \mathcal{G}_0$, then $||\widehat{\Upsilon}(x) - \widehat{\Upsilon}(x')||^2_2  \leq 1 - k(x, x')$ if $|\mathcal{G}_0|=1$. Further, if $|\mathcal{G}_0|>1$ then $||\widehat{\Upsilon}(x) - \widehat{\Upsilon}(x')||^2_2  < 1 - k(x, x')_H$ for all $x, x' \in \mathcal{X}$. Here $k(x, x')_H$ is the kernel distance in the Hausdorff sense over $\mathcal{G}_0$ in $\mathcal{H}$, \emph{i.e.} $k(x, x')_H = \max_{g, g' \in\mathcal{G}_0\subseteq \mathcal{G}}k(gx, g'x'). $
\end{theorem}
Thus $\widehat{\Upsilon}(x)$ can achieve partial invariance provided a limited number of transformations of the unlabelled data. Further, results developed for kernel methods in this section encourage their use in practice since the feature $\widehat{\Upsilon}(x)$ now solves both motivating problems mentioned in Section~\ref{section_1}. Note that \textit{the notion of and results on partial invariance can be easily applied to the invariant kernel $k_\Psi(x, y)$ proposed in Section~\ref{sec_group_invariance} and \ref{sec_inv_kernel} thereby making them practical tools with theoretical guarantees. }


\begin{table}[!t]
\small
\caption{Mean 10-fold cross validation testing classification accuracy (\%), of a linear SVM tested on different features of  $\mathcal{X}_{\mathcal{G}_0Te}$ while it was trained on the corresponding features of $\mathcal{X}_{Tr}$.} 
\centering 
\begin{tabular}{| c | c || c | c c c |} 
\hline 
Dataset & Raw $\mathcal{X}_{Te}$ & Raw $\mathcal{X}_{\mathcal{G}_0Te}$ & $\mu^k_n(\mathcal{X}_{\mathcal{G}_0Te})$ & $\widehat{\Upsilon}(\mathcal{X}_{\mathcal{G}_0Te})_{RBF}$ & $\widehat{\Upsilon}(\mathcal{X}_{\mathcal{G}_0Te})_{poly}$ \\
\hline 
 banana & 55.2 & 55.2 & 60.56 & \textbf{61.70} & 60.37 \\ 
breast & 71.34 & 64.77 & 71.56 & 70.42 & \textbf{71.58} \\
german & 76.03 & 62.48 & 69.63 & \textbf{69.78} & 69.63 \\ 
diebetis & 75.67 & 50.35 & 65.84 & \textbf{66.20} & 65.84 \\ 
image & 83.81 & 57.05 & 57.62 & \textbf{60.27} & 57.49 \\
splice & 84.54 & 55.03 & 55.07 & \textbf{79.83} & 55.07 \\
thyroid & 91.53 & 52.67 & 66.76 & 64.63 & \textbf{68.46} \\
 ringnorm & 77.26 & 43.68 & \textbf{57.67} & 56.48 & 56.85 \\ 
 twonorm & 97.59 & 31.48 & \textbf{69.21} & 66.06 & 64.69 \\
waveform & 89.87 & 63.77 & 65.50 & 64.64 & \textbf{66.89} \\

\hline 

\end{tabular}
\label{tab_exp_1} 
\end{table}

\section{Experimental Validation}\label{sec_exp}

\textbf{Goal:} The goal of this section is two fold, to see (1) whether partially invariant kernel features $\widehat{\Upsilon}(x)$ and (2) invariant kernel SVM \emph{i.e.} objective~\ref{eq_svm} coupled with Theorem~\ref{invariant_kernel}, in practice, are able to address Problem 1 and Problem 2, and (3) whether kernel invariant features offer any advantage over linear invariant features $\mu(x)$. We refrain from using discriminative kernel features since our theoretical results does not assume any structure for the templates.

\textbf{Set-up and Method:} We use 10 normalized datasets (each with number of samples $ \geq 200$) from the UCI ML repository for this task.  We form a random 10-fold cross-validation partition (training ($\mathcal{X}_{Tr}$)/testing ($\mathcal{X}_{Te}$)) for each dataset $\mathcal{X}$. In order to enforce Problem 1, we introduce a number of transformations $g$ belonging the a randomly chosen set of unitary transformations $\mathcal{G}_0$ into the test data ($\mathcal{X}_{Te}$) thereby multiplying the test data size by a factor of $|\mathcal{G}_0|$, thus obtaining $\mathcal{X}_{\mathcal{G}_0Te}$ (we set $|\mathcal{G}_0|=10$)\footnote{We uniformly set $|\mathcal{G}_0|$ to a reasonably modest value of 10 in order to keep computational load of multiplying the dataset manageable.}. However, we do not augment the training data $\mathcal{X}_{Tr}$. We instead generate random vectors or templates $t\in T$ and augment them using the same unitary transformations $\mathcal{G}_0$ as the test data (we set $|T|=100$ for all experiments). This enforces Problem 1. Problem 2 is inherently enforced to a large degree since it is practically  very difficult to generate an entire group. The transformations we introduce are a subset of the unitary group \emph{i.e.} $\mathcal{G}_0 \subset \mathcal{G}$ with $|\mathcal{G}_0|=10$.

\begin{table}
\small
\caption{Mean 10-fold cross validation testing classification accuracy (\%). $\mathcal{X}_{Te}$ and S.K $\mathcal{X}_{\mathcal{G}_0Te}$ denote results for a standard kernel and I.K $\mathcal{X}_{\mathcal{G}_0Te}$ denotes it for the invariant kernel (Theorem~\ref{invariant_kernel}).}\label{tab_exp_2} 
\centering 
\begin{tabular}{| c | c || c | c |} 
\hline 

Dataset & $\mathcal{X}_{Te}$ & S.K $\mathcal{X}_{\mathcal{G}_0Te}$ & I.K $\mathcal{X}_{\mathcal{G}_0Te}$  \\
\hline 
 banana & 72.34 & 47.16 & \textbf{50.67}  \\
 breast & 66.15 & 62.31 & \textbf{67.27}  \\
 german & 70.60 & 70.00 & 70.07  \\
 diabetis & 73.68 & 44.67 & \textbf{65.39}  \\
 image & 95.96 & 43.08 & \textbf{56.34}  \\
 splice & 58.93 & 55.08 & 55.08  \\
 thyroid & 90.48 & \textbf{64.81} & 64.57  \\
 ringnorm & 69.16 & 43.83 & \textbf{50.46}  \\
 twonorm & 97.11 & 32.66 & \textbf{49.96}  \\
 waveform & 72.86 & 67.06 & 67.06  \\

\hline 

\end{tabular}
\label{tab_exp_2} 
\end{table}

For our first experiment, we compute $\widehat{\Upsilon}(x)$ using the randomly generated transformed templates $T_{\mathcal{G}_0}$ and use the RBF kernel ($\sigma = 1$) and the polynomial kernel ($k(x, y) = (\langle x, y \rangle + 1)^d$ with degree $d = 2$). We set $\eta$ to compute the infinite moment equivalent to max-pooling. As an evaluation, to estimate the separability of the data, we train a linear SVM on the unaugmented (not transformed) data ($\mathcal{X}_{Tr}$) using (1) raw features (Raw baseline), (2) linear invariant features ($\mu^k_n(x)$ baseline), (3) RBF kernel invariant features ($\widehat{\Upsilon}(x)_{RBF}$) and (4) polynomial kernel invariant features ($\widehat{\Upsilon}(x)_{poly}$). We then test on the augmented corresponding fold (transformed) of the test data ($\mathcal{X}_{\mathcal{G}_0Te}$) after extracting the corresponding feature. We also report the test accuracy of testing on raw $\mathcal{X}_{Te}$ as an illustration of the classification difficulty introduced by the transformations added in. The results are summarized in Table~\ref{tab_exp_1}. For our second experiment, we use the same datasets and generate a random 10-fold partition. Here we always train on the \textit{raw} untransformed ($\mathcal{X}_{Tr}$) fold and test on the raw \textit{transformed} data  ($\mathcal{X}_{\mathcal{G}_0Te}$).
We train a standard RBF kernel SVM ($\sigma=1$) and an invariant SVM using the same kernel as described in Theorem~\ref{invariant_kernel}. We also test the standard kernel SVM on the untransformed data ($\mathcal{X}_{Te}$) as an illustration of the classification difficulty introduced due to transformations. The results are summarized in Table~\ref{tab_exp_2}.

\textbf{Results:} Our first observation is that in almost all of the datasets, even the modestly ($|\mathcal{G}_0|=10$) added transformations significantly impaired the SVM's performance (Table~\ref{tab_exp_1} and Table~\ref{tab_exp_2}). Thus we confirm that most of the difficulty in the problem of learning arises from the presence of inherent transformations relating different orbits of the data. Secondly, for both experiments explicitly generating invariance through invariant features (Table~\ref{tab_exp_1}) and through the invariant kernel (Table~\ref{tab_exp_2}) helps in performance suggesting that in both cases sample complexity was lowered. We find that invariant kernel features and the invariant kernel (Theorem~\ref{invariant_kernel}), in practice as well, address Problem 1 and Problem 2. Kernel features, in general, modestly outperform linear features in most of these datasets since even though the features are non-linear, the transformation they are invariant to are linear.

\section{Conclusion}

One of the main handicaps in applying invariant kernel methods was the computational expense in generating and processing additional transformed form of the data. Further, in many cases it is difficult to generate such samples due to the transformation being unknown. However, in many cases, it is easier to obtain transformed unlabelled samples (such as video sequences in vision). The invariant kernels described in this paper can be used to address such issues while theoretically guaranteeing invariance.


\bibliography{iclr2016_conference}
\bibliographystyle{iclr2016_conference}

\newpage

\section{Supplementary Material}

\subsection{Proof of Lemma~\ref{lem_invariance}}
\begin{proof}We have,
$$g' \int_\mathcal{G} g \omega ~dg = \int_\mathcal{G} g'g \omega ~dg =  \int_\mathcal{G} g'' \omega ~dg'' = \int_\mathcal{G} g \omega ~dg$$

Since the normalized Haar measure is invariant, \emph{i.e.} $dg = dg'$. Intuitively, $g'$ simply rearranges the group integral owing to elementary group properties.
\end{proof}

\subsection{Proof of Lemma~\ref{transpose}}

\begin{proof} We have,
$$\Psi^T = (\int_\mathcal{G} g~dg)^T = \int_\mathcal{G} g^T~dg = \int_\mathcal{G} g^{-1}~dg^{-1} = \Psi$$
Using the fact $g\in \mathcal{G} \Rightarrow g^{-1} \in \mathcal{G}$ and $dg = dg^{-1}$.
\end{proof}

\subsection{Proof Lemma~\ref{unitary_projection}}

\begin{proof} We have,
$$\Psi \Psi = \int_\mathcal{G}\int_\mathcal{G} gh ~dg ~dh =  \int_\mathcal{G}\int_\mathcal{G} g' dg' dh =  \int_\mathcal{G} dh \int_\mathcal{G} g' dg' = \Psi $$
Since the Haar measure is normalized ($\int_\mathcal{G} dg = 1$), and invariant. Also for any $\omega, \omega' \in \mathbb{R}^d$, we have 
$\langle \omega, \Psi \omega' \rangle = \int_\mathcal{G}\langle \omega,  g\omega' \rangle dg = \int_\mathcal{G}\langle g^{-1}\omega,  \omega' \rangle dg^{-1} = \langle \Psi \omega, \omega'  \rangle$
\end{proof}

\subsection{Proof of Lemma~\ref{invariant_proj}}
\begin{proof}
We have $ \langle \Psi \omega, \Psi\omega' \rangle = \langle\int_{g} g \omega, \Psi \omega'\rangle dg =  \langle\int_{g} g' \omega, \Psi \omega'\rangle dg = \langle g'\omega, \Psi\omega') \int_{g} dg = \langle g'\omega, \Psi\omega'\rangle$

In the second equality, we fix a group element $g'$ since the inner-product is invariant using the argument $ \langle \phi\omega ,\Psi\omega'\rangle = \langle g'\omega, \Psi\omega'\rangle$. This is true using Lemma~\ref{lem_invariance} and the fact that $\mathcal{G}$ is unitary. Further, the final equality utilizes the fact that the Haar measure $dg$ is normalized.
\end{proof}


\subsection{Proof of Theorem~\ref{theorem_1}}
\begin{proof}
We have $\langle \phi(gx), \phi(gy)\rangle = \langle \phi(x), \phi(y)\rangle = \langle g_\mathcal{H} \phi(x), g_\mathcal{H} \phi(y)\rangle$, since the kernel $k$ is unitary. Here we define $g_\mathcal{H} \phi(x)$ as the action of $g_\mathcal{H} $ on $\phi(x)$. Thus, the mapping $g_\mathcal{H}$ preserves the dot-product in $\mathcal{H}$ while reciprocating the action of $g$. This is one of the requirements of a unitary operator, however $g_\mathcal{H}$ needs to be linear. We note that linearity of $g_\mathcal{H} $ can be derived from the linearity of the inner product and its preservation under $g_\mathcal{H}$ in $\mathcal{H}$. Specifically for an arbitrary vector $p$ and a scalar $\alpha$, we have 
\begin{align}
|| \alpha g_\mathcal{H}p -  g_\mathcal{H} (\alpha p)||^2 &= \langle  \alpha g_\mathcal{H}p -  g_\mathcal{H} (\alpha p),  \alpha g_\mathcal{H}p -  g_\mathcal{H} (\alpha p)  \rangle\\
&= ||\alpha g_\mathcal{H}p ||^2 + || g_\mathcal{H} (\alpha p)||^2  - 2 \langle \alpha g_\mathcal{H}p ,  g_\mathcal{H} (\alpha p)  \rangle\\
&=  |\alpha|||p ||^2 + || \alpha p||^2  - 2 \alpha^2 \langle  p ,p  \rangle = 0\\
\end{align}
Similarly for vectors $p, q$, we have $ || g_\mathcal{H}(p+q) -  (g_\mathcal{H} p +g_\mathcal{H} q)||^2 = 0$

We now prove that the set $\mathcal{G}_\mathcal{H}$ is a group. We start with proving the closure property. We have for any fixed $g_\mathcal{H}, g'_\mathcal{H}\in \mathcal{G}_\mathcal{H}$
$$ g_\mathcal{H}g'_\mathcal{H}\phi(x) = g_\mathcal{H}\phi(g'x) = \phi(gg'x) = \phi(g''x)  = 
g''_\mathcal{H}\phi(x)$$

Since $g''\in \mathcal{G}$ therefore $g''_\mathcal{H}\in \mathcal{G}_\mathcal{H}$ by definition. Also, $ g_\mathcal{H}g'_\mathcal{H} =  g''_\mathcal{H}$ and thus closure is established. Associativity, identity and inverse properties can be proved similarly. The set $\mathcal{G}_\mathcal{H} = \{ g_\mathcal{H} ~|~ g_\mathcal{H}: \phi(x)\rightarrow \phi(gx)~\forall g\in \mathcal{G} \}$ is therefore a unitary-group in $\mathcal{H}$.
\end{proof}

\subsection{Proof of Theorem~\ref{th_generalization}}
\begin{proof} Since $\Psi_\mathcal{H}\omega^*$ is a perfect separator for $\{\Psi_\mathcal{H}\phi(\mathcal{X})\}$, $\exists \rho' > 0$, s.t. $\min_i y_i (\Psi_\mathcal{H}\phi(x_i))^T(\Psi_\mathcal{H}\omega^*) \geq \rho'~\forall \{ x_i, y_i\}\in \mathcal{X}$.

Using Lemma 2.4 and Theorem~\ref{theorem_1}, we have for any fixed $g'_\mathcal{H} \in \mathcal{G}_\mathcal{H}$, 
$$ (\Psi_\mathcal{H} \phi(x_i))^T(\Psi_\mathcal{H}\omega^*) = ( g'_\mathcal{H} \phi(x_i))^T(\Psi_\mathcal{H}\omega^*)$$


 Hence, $$ \min_i y_i ( g'_\mathcal{H} \phi(x_i))^T(\Psi_\mathcal{H}\omega^*) = \min_i y_i (\Psi_\mathcal{H}\phi(x_i))^T(\Psi_\mathcal{H}\omega^*) \geq \rho'~~\forall (g'_\mathcal{H} \Rightarrow g)\in \mathcal{G}$$
Thus, $\Psi_\mathcal{H}\omega^*$ is perfect separator for $\{\phi(\mathcal{X}_\mathcal{G})\}$ with a margin of at-least $\rho'$. It also implies that a max-margin separator of $\{\Psi_\mathcal{H}\phi(\mathcal{X})\}$ is also a max-margin separator of $\{\phi(\mathcal{X}_G)\}$.
\end{proof}

\subsection{Proof of Theorem~\ref{invariant_kernel}}
\begin{proof}
For any fixed $g'_\mathcal{H}$ we find, $ \langle \Psi_\mathcal{H} \phi(x), \Psi_\mathcal{H}\phi(y) \rangle = \langle g'_\mathcal{H}\phi(x), \Psi_\mathcal{H}\phi(y) \rangle$ using Lemma~\ref{unitary_projection}. Choosing $g'_\mathcal{H}$ to be identity and substituting the expansion of $\Psi_\mathcal{H}$, $\phi(x)= \phi(T)u_x$ and  $\phi(y)=\phi(T)u_y$ we have the desired result.
\end{proof}

\subsection{Proof of Theorem~\ref{theorem_stability}}
\begin{proof}
Since $\eta_n$ are Lipschitz continuous $\forall n$, for each $k$ component of the signature $\Upsilon^k_n(x)$, we have
\begin{align}
||\Upsilon^k(x) - \Upsilon^k(x')||^2_{\mathbb{R}^N} &\leq \frac{1}{|\mathcal{G}|^2} \sum_n\left( \sum_{g\in\mathcal{G}_\mathcal{H}} L_{\eta_n}|\langle \phi(x), g_\mathcal{H}\omega_g^k \rangle - \langle \phi(x'), g_\mathcal{H}\omega_g^k  \rangle | \right)^2\\
& \leq \frac{L^2_\eta}{|\mathcal{G}|^2} \sum_n  \left(\sum_{g\in\mathcal{G}_\mathcal{H}} ||\phi(x)-\phi(x')||_\mathcal{H} ||g_\mathcal{H}\omega_g^k ||_\mathcal{H} \right)^2 \\
& \leq N^2 L^2_\eta ||\phi(x) - \phi(x')||_\mathcal{H}
\end{align}

where we utilize Cauchy-Schwartz, Theorem~\ref{theorem_1} and the fact that for some $t^k\in \mathbb{R}^d$, we have $||\omega_g^k ||_2^2 = ||\phi(t^k)||_2^2 = \langle \phi(t^k), \phi(t^k) \rangle = k(t^k, t^k) = 1$. Since $\Upsilon^k(x)$ is invariant to the action of $\mathcal{G}$ (and consequently $\mathcal{G}_\mathcal{H}$), $||\Upsilon^k(x) - \Upsilon^k(x')||_{\mathbb{R}^N}\leq N^2L^2_\eta \min_{g_\mathcal{H}, g'_\mathcal{H}\in \mathcal{G}_\mathcal{H}} || g_\mathcal{H}\phi(x) - g'_\mathcal{H}\phi(x') ||^2_\mathcal{H} = N^2L^2_\eta \min_{g, g'\in \mathcal{G}} || \phi(gx) - \phi(g'x') ||^2_\mathcal{H} =  N^2L^2_\eta||\phi(x)-\phi(x')||^2_H = 2N^2L^2_\eta(1- k(x, x')_H)$. If $NL_\eta <\frac{1}{\sqrt{2}}$, then the map is a contraction and we obtain the desired result by summing over all $K$ components and dividing by $K$.
\end{proof}

\subsection{Proof of Theorem~\ref{theorem_partial_inv}}
\begin{proof}
The proof is very similar to that of Theorem 6 in \cite{AnselmiLRMTP13} since Theorem~\ref{theorem_1} allows the group structure of $\mathcal{G}_\mathcal{H}$ to be preserved in $\mathcal{H}$. 
\end{proof}

\subsection{Proof of Theorem~\ref{theorem_stability_partial}}
\begin{proof}
The first condition follows the exact analysis as in Theorem~\ref{theorem_stability}. For the second condition to hold, we apply Theorem 5.1 and follow an argument very similar to that of Theorem~\ref{theorem_stability}.
\end{proof}

\end{document}